\documentclass[letterpaper, 10 pt, conference]{IEEEtran}
\IEEEoverridecommandlockouts 
\usepackage{cite}
\usepackage{graphics} 
\usepackage{epsfig} 
\usepackage{mathptmx} 
\usepackage{times} 
\usepackage{amssymb} 
\usepackage[cmex10]{amsmath}
\usepackage{siunitx}
\usepackage{amsthm}
\usepackage{amsfonts}
\usepackage{latexsym}
\usepackage{mathrsfs} %
\usepackage{xcolor} %
\usepackage[normalem]{ulem}
\usepackage{float}
\usepackage{mathtools}
\usepackage{wrapfig}
\usepackage{xspace}
\usepackage{pbox}
\usepackage{geometry} %
\usepackage{graphicx} %
\usepackage{hybrid} %
\usepackage{caption}
\usepackage{flushend}
\ifCLASSOPTIONcompsoc
    \usepackage[caption=false, font=normalsize, labelfont=sf, textfont=sf]{subfig}
\else
\usepackage[caption=false, font=footnotesize]{subfig}
\fi

\usepackage[linkbordercolor={1 1 1},citebordercolor={1 1
  1},urlbordercolor={0.0 0.0
  0.0},urlcolor=blue,colorlinks=true,linkcolor=black,citecolor=black]{hyperref}
\usepackage{url}

\title{\LARGE \bf \vspace{18pt} Improved Performance on Moving-Mass Hopping Robots with Parallel Elasticity}

\author{Eric Ambrose and Aaron D. Ames%

\thanks{*This work was supported by Disney Research and Development.}%
\thanks{Eric Ambrose and Aaron Ames are with the Department of Mechanical and Civil Engineering, California Institute of Technology, Pasadena, CA 91125. }%
}

\begin{document}

\maketitle
\thispagestyle{empty}
\pagestyle{empty}

\begin{abstract}

Robotic Hopping is challenging from the perspective of both modeling the dynamics as well as the mechanical design due to the short period of ground contact in which to actuate on the world. Previous work has demonstrated stable hopping on a moving-mass robot, wherein a single spring was utilized below the body of the robot. This paper finds that the addition of a spring in parallel to the actuator greatly improves the performance of moving mass hopping robots. This is demonstrated through the design of a novel one-dimensional hopping robot. For this robot, a rigorous trajectory optimization method is developed using hybrid systems models with experimentally tuned parameters. Simulation results are used to study the effects of a parallel spring on energetic efficiency, stability and hopping effort. We find that the double-spring model had 2.5x better energy efficiency than the single-spring model, and was able to hop using 40\% less peak force from the actuator. Furthermore, the double-spring model produces stable hopping without the need for stabilizing controllers. These concepts are demonstrated experimentally on a novel hopping robot, wherein hop heights up to 40cm were achieved. 

\end{abstract}

\section{Introduction}
\label{sec:intro}

Hopping is a mode of locomotion that undergoes rapid changes in contact with the world; as a result, advanced methods from nonlinear and hybrid dynamics are well-suited for its analysis \cite{sinnet2011human}, \cite{westervelt2003}. Just as in the case of running with legged robots \cite{ma2017bipedal}, \cite{raptor}, hopping consists of long airborne phases and shorter ground contact phases separated by dramatic discrete impacts. These impacts, along with friction and other forces, cause energy to be lost at every hop. This necessitates adding energy back into the system through its actuators in order to reach the desired hop height in a controlled manner. The most straight forward way to accomplish this is to apply force on the world during the ground contact phase to assist the robot's ascent. Due to the limited time on the ground, this leads to a need for brief and high power actuation coupled with compliant elements to store and return energy.

Over the past few decades, robotic hoppers have been developed with many approaches to the challenges of design and control \cite{RaibertHop2D}, \cite{RaibertHop3D}, \cite{Fiorini2003}. These robots successfully hop for long periods of time, avoid large obstacles, and perform extreme behaviors such as flips \cite{LeggedBalance}. Beyond this, they helped set the standard for the level of control to be expected for such a dynamic type of locomotion. One of the main focuses of hopping robot design has been to improve the efficiency of locomotion \cite{bowleg2D}, \cite{Capler-Leg}, \cite{salto} to allow for long-term hopping using on-board power. These improvements were achieved through a variety of means such as lessening the mass of components, using more direct actuation input, or using more efficient elastic elements. With improved efficiency of actuation, researchers have been able to push the capabilities of the robots further towards higher hops \cite{Capler-Leg}, \cite{salto} and over complex terrain \cite{hubickiHop}.

\begin{figure}[t!]
	\centering
	\includegraphics[width=0.95\columnwidth]{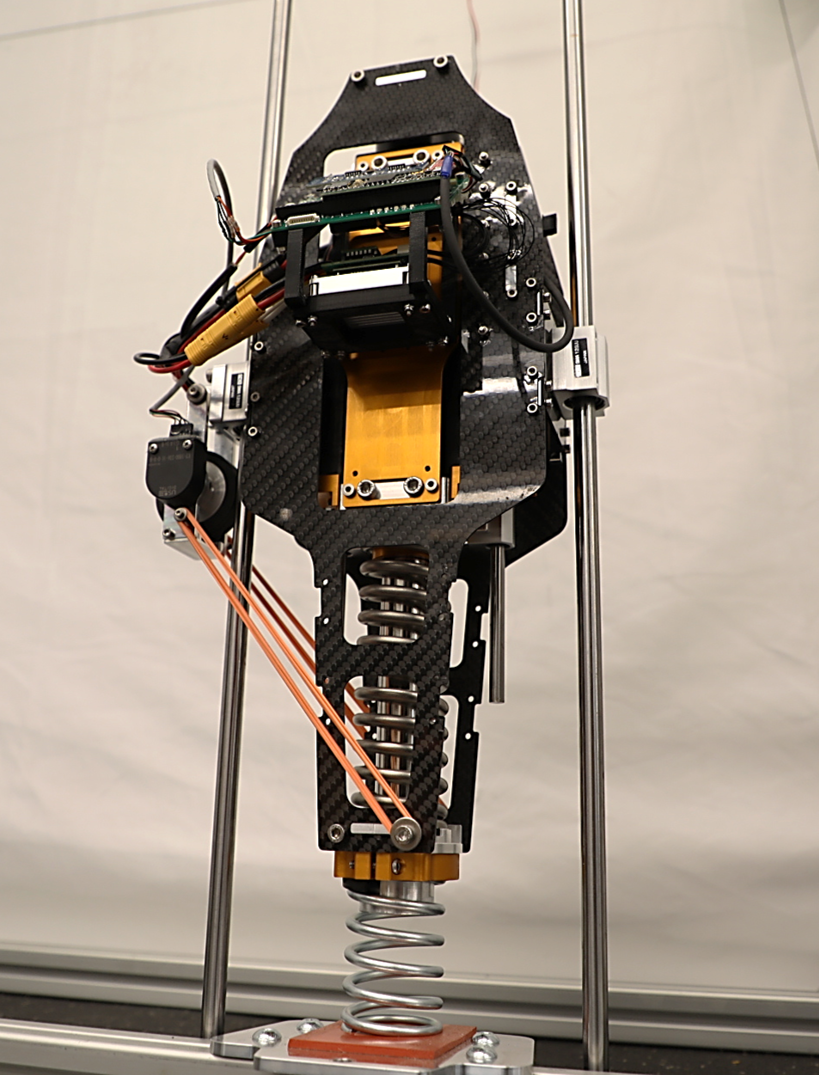}
	\setlength\belowcaptionskip{2pt}
	\caption{A 3-mass, vertically-constrained hopping robot with two springs: one in parallel with the actuator and one below in series with the actuator.}
	\label{pixar}
\end{figure}

When using high-power actuation in critical environments such as when humans are present, the task of safety becomes a critical concern. The issue of robotic safety has been brought up in the field of robotic manipulators, due to the more recent prominence of manufacturing and surgical robots. Measures have been to taken to deal with this issue in a variety of ways, including the addition of elasticity \cite{safeArm_Yoon}, \cite{lensSafeHRI}. In the realm of robotic hopping, safety can be improved by placing a spring between the robot and the world, as was done with the LEAP robot developed by Disney Research and Development \cite{url:leap}. The goal of putting an elastic element here is to limit the impulse force transferred from the robot to prevent any possible damage or harm on the surroundings. This can be further seen in the patent of a robotic bouncing ball \cite{bouncingBallPatent}, where all actuators and mechanisms are contained within the ball itself.

With the goal of finding robotic hopping solutions that fit within the paradigm of this safe actuation, previous research \cite{iros19} examined two methods of hopping: a clutch-release hopper similar to that of the Bowleg Hopper \cite{bowleg2D}, and a moving-mass hopper akin to that used in \cite{aguilar2016Jump}. It was shown that both types of hopping could reach a stable desired hop height through control methods in the span of a few hops. The previous model for the moving-mass hopper had a single spring below the body of the robot which would act as a safe barrier between it and the world, while also providing the necessary energy storage to achieve periodic hopping. The results with this model showed a need for large actuation force in order to reach a desired hop height.

This work focuses on improving the design of the moving-mass hopper model through the addition of a second spring, placed in parallel with the actuator. Research groups have shown that there can be an improvement in efficiency to robot locomotion with the incorporation of parallel elasticity \cite{PEA_efficient}, \cite{uPEA_Effects}. Furthermore, it has been shown that parallel elastic actuators can be used to improve the tracking of motion trajectories \cite{PEA_traj}, such as those used in the original moving-mass model. A new robot was created for this work, with the addition of the second spring in parallel in order to examine the benefits of parallel elasticity. Section \ref{sec:model} will go over the new model and the structure of its corresponding hybrid control system. Section \ref{sec:sim} will explain the motion generation and simulation methods used to analyze the model. Then in Section \ref{sec:exp}, the hopping robot will be introduced along with the experimental results. Finally, the conclusions will be presented in Section \ref{sec:conclusion}, discussing how higher hopping was achieved with better energetic efficiency and stability properties.            
\section{Hopping Model}
\label{sec:model}

\begin{figure}[t!]
	\centering
	\includegraphics[width=0.95\columnwidth]{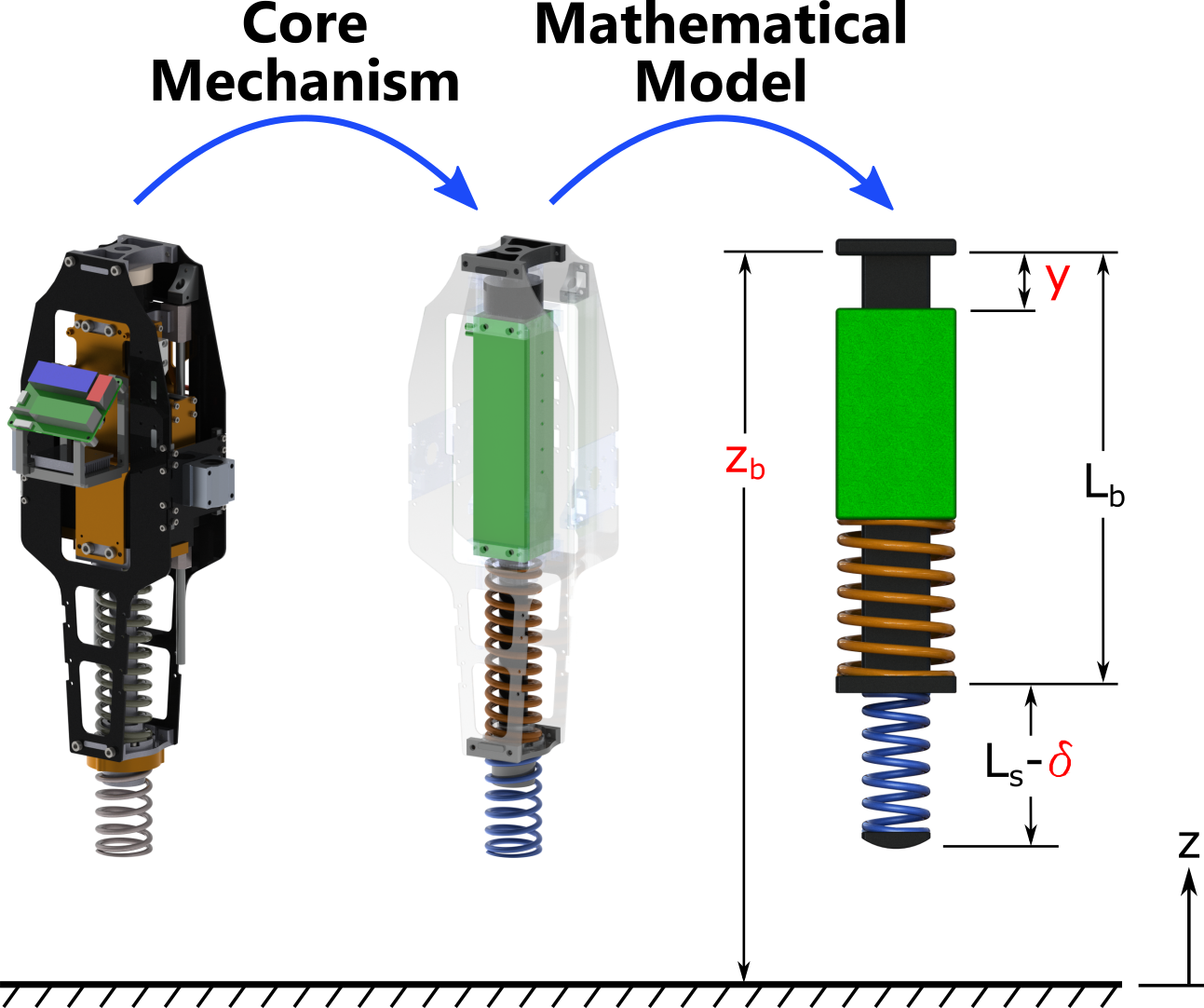}
	\setlength\belowcaptionskip{6pt}
	\caption{Model coordinates and design for the double-spring, moving-mass robot, with a look at the core mechanism.}
	\label{model3_config}
\end{figure}

The original moving-mass model was made up of three masses and a single spring, acting between the body and foot masses. The third mass, the mover, was contained within the body and could be moved vertically with the use of linear actuation. This arrangement had the spring in series with the actuator which would serve to redirect the motion of the body back upwards after an impact with the ground. The actuator could slide the mover impulsively while the robot was on the ground in order to assist the reversal of the body's velocity. The actuator also needed to apply force in order to keep the mover from reaching the end of its range of motion and creating an internal impact which could reduce the energy in the system and lead to an unwanted loss of hop height.

Unlike this previous model, the new model presented here contains an additional spring between the body and mover which acts in parallel with the actuator. By adding this second spring here, the mover will now undergo its own passive reversal of velocity during the ground phase. A caveat of this design, is that the mover will now contact the body when the spring is at its equilibrium length, meaning that the spring will only compress and not extend past this point. This leads to internal contact occurring between the body and mover at the end of the second springs compression cycle.

\subsection{Dynamics}
\label{subsec:dynamics}

In order to model this new double-spring hopping robot, the coordinates of the configuration space are chosen to be $q=(z_{b},y,\delta) \in Q \subset \R^{3}$, where $z_{b}$ is the height of the body from the ground, $y$ is the deflection of the upper spring in parallel, and $\delta$ is the deflection of the lower spring in series. These are shown in red within Fig. \ref{model3_config}, along with two of the constant length parameters of the robot model. The unpinned dynamics were derived using these coordinates and Lagrangian mechanics, then put in the form

\begin{equation}\label{eq:EoM3}
	\underbrace{
		\begin{bmatrix}
			M_{0} & -M_{m} & M_{f} \\
			-M_{m} & M_{m} & 0 \\
			M_{f} & 0 & M_{f} \\
		\end{bmatrix}
	}_{M}
	\underbrace{
		\begin{bmatrix}
		    \ddot{z}_{b} \\ \ddot{y} \\ \ddot{\delta} \\
		\end{bmatrix}
	}_{\ddot{q}}
	+
	\underbrace{
		\begin{bmatrix}
    		c_{b}\dot{z}_{b} + M_{0}g \\
    		\textcolor{red}{F_{p}} - M_{m}g \\
    		F_{s} + M_{f}g \\
		\end{bmatrix}
	}_{H(q,\dot{q})}
	=
	\underbrace{
		\begin{bmatrix}
		    0 \\ 1 \\ 0 \\
		\end{bmatrix}
	}_{B}
	u
\end{equation}
where $M_{0}$ is the total mass of the robot, $M_{m}$ is the mass of the mover, $M_{f}$ is the mass of the foot, $C_{b}$ is the damping coefficient of the body moving vertically, $F_{p}$ is the extension force in the upper spring in parallel, $F_{s}$ is the extension force of the lower spring in series, and $g$ is the gravitational acceleration constant. The full form of the spring forces includes a term of the stiffness force and a term for the damping force, i.e. $F_{p}=k_{p}y+c_{p}\dot{y}$, where $k_{p}$ and $c_{p}$ are the stiffness and damping constants of the spring in parallel.

The sole difference in the unpinned dynamics of this robot model from the single-spring case comes from the $F_{p}$ term, which is highlighted in (\ref{eq:EoM3}). For the case of the single-spring model, this term was just the damping force between the mover and body. While this is a small difference in the unpinned dynamics, the complexity from the internal contact is more significant, as will be shown in the following subsection.

\subsection{Hybrid Structure}
\label{subsec:hybrid}

\begin{figure}[t!]
	\centering
	\includegraphics[width=0.9\columnwidth]{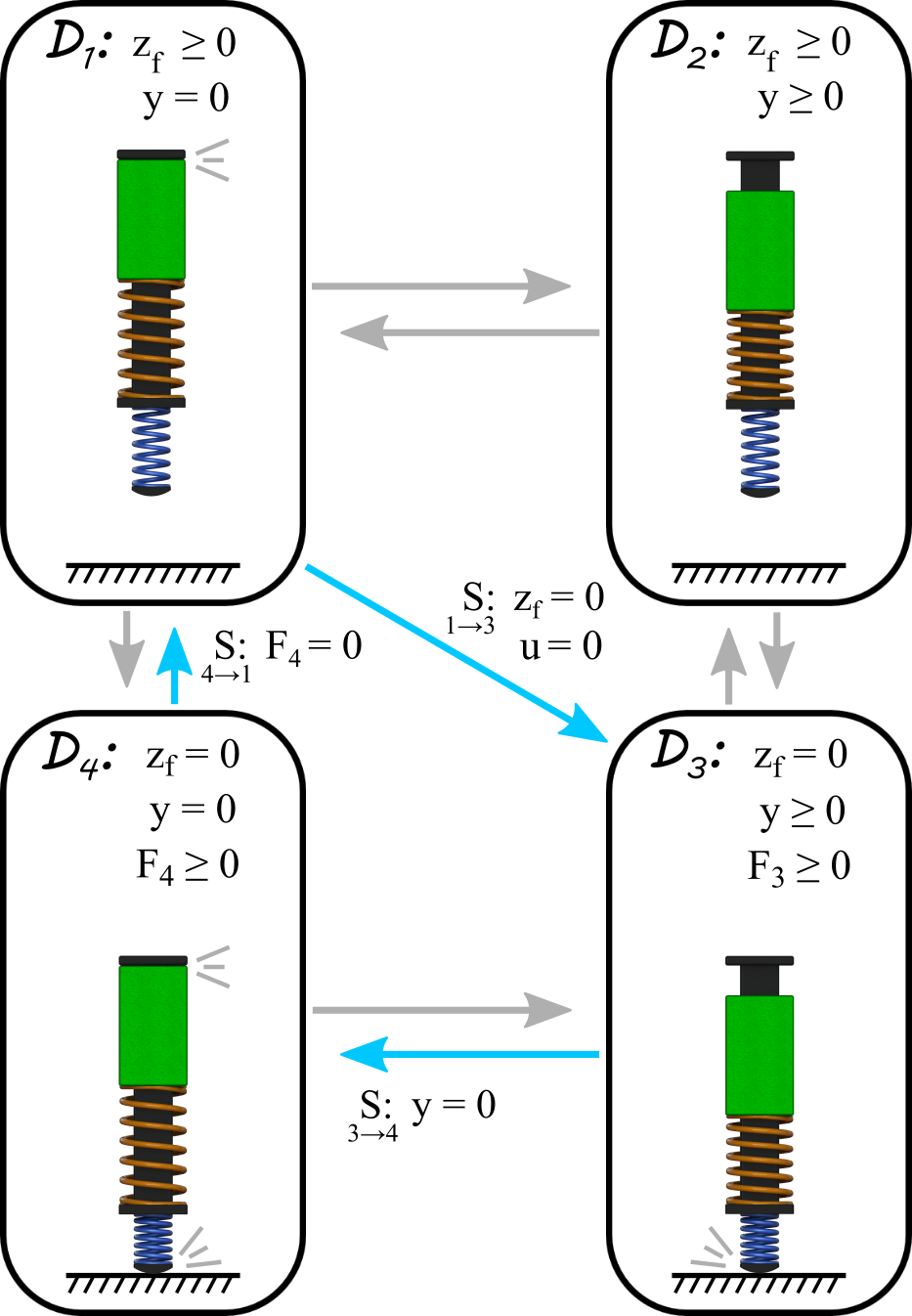}
	\setlength\belowcaptionskip{2pt}
	\caption{Directed cycle for the hybrid system; Vertices depicted as boxes and edges shown as arrows. Blue arrows show the specific domain map for this work, and gray arrows show other possible switches.}
	\label{model3_hybrid}
\end{figure}

Beyond the dynamics described in (\ref{eq:EoM3}), there are also external and internal forces acting on the robot intermittently during each hop, which come from the contact with the ground, as well as a range of motion limit placed around the spring in parallel within the body. This hardstop prevents the upper spring from extending past its equilibrium length. The mover is able to move downwards within the body, compressing the upper spring, but will impact the body when it reaches the equilibrium position again. At the point of some of these discrete events an impact occurs, such as when the foot reaches the ground. Due to these changes in the dynamics and the presence of discrete impact events, the dynamical model can be re-written as a hybrid control system. For the tangent bundle with coordinates $(q,\dot{q}) \in TQ \subset \R^{6}$, the hybrid control system is defined as the tuple,
\begin{equation}
	\label{eq:hybrid-control-system}
	\HybridControlSystem = (\DirectedGraph,\Domain,\ControlInput,\Guard,\ResetMap,\emph{FG})
\end{equation}
\begin{itemize}
	\item $\DirectedGraph = \{\Vertex,\Edge\}$ is a \emph{directed cycle} with vertices $\Vertex$ and edges $\Edge$. Fig. \ref{model3_hybrid} shows the cycle for the double-spring model
	
	\item $\ControlInput \in \R$ is the set of admissible control input forces
	
	\item $\Domain=\{\Domain_{v} \}_{v \in V}$ is the set of admissible domains
	
	\item $\Guard=\{\Guard_{\ei}\}_{\ei \in  E}$ is the set of guards for domains, which represents the transition conditions between domains
	
	\item $\ResetMap = \{\ResetMap_{\ei}\}_{\ei \in  E}$ is the set of state reset maps between domains, defined in (\ref{eq:impact})
	
	\item $\emph{FG}$ is the set of vector fields representing the control system dynamics for each domain, defined in (\ref{eq:dyn})-(\ref{eq:Fv})
\end{itemize}

For the case of this double-spring hopper model, there are four possible domains for the robot as shown in Fig. \ref{model3_hybrid}. These are based on the state of ground contact and internal hardstop contact, which are visually represented in the figure by the gray animation effect lines. The arrows in the figure represent the possible domain switching that can occur. Note that there is only one  diagonal arrow in the diagram. This is due to an assumption that two contact switching events will not occur at the same time, with the exception of if the robot hits the ground while applying zero actuator force, in which case it will transition straight from domain $1$ to domain $3$. Each contact manifests in the dynamics through a holonomic constraint, which can be added onto the end of our equations of motion to yield the new set of equations augmenting (\ref{eq:EoM3}):
\begin{equation} \label{eq:dyn}
    M\ddot{q}+H(q,\dot{q})=Bu+J_{v}^{T}F_{v}
\end{equation}
\begin{equation} \label{eq:Jacobian}
    J_{v}\ddot{q} + \dot{J}_{v}\dot{q}=0
\end{equation}
where $J_{v}$ is the Jacobian of the holonomic constraints in domain, $D_{v}$, and $F_{v}$ are the forces of those same constraints. Domain $D_{2}$ does not include any contacts, unlike the other three domains. The Jacobians for the contact domains are:
\begin{equation}\label{eq:jacobi}
    J_{1} =
    	\begin{bmatrix}
    	    0 & 1 & 0 \\
    	\end{bmatrix},\;
    J_{3} =
    	\begin{bmatrix}
    	    1 & 0 & 1 \\
    	\end{bmatrix},\;
    J_{4} =
        \begin{bmatrix}
    	    0 & 1 & 0 \\
    	    1 & 0 & 1 \\
    	\end{bmatrix}
\end{equation}

The method of defining these Jacobians and their forces is based on the concepts presented in \cite{MLS94} where the constraint force, $F_{v}$, is found using the following calculation:
\begin{equation}\label{eq:Fv}
F_{v}(q,\dot{q})=-(J_{v}M^{-1}J_{v}^{T})^{-1} [J_{v}M^{-1}(Bu-H(q,\dot{q}))+\dot{J}_{v}\dot{q}]
\end{equation}

As mentioned before, there are impacts which occur at some of the domain transitions. This includes any time that the foot contacts the ground and anytime that the mover reaches its range of motion limit against the body of the hopper. The impact in each case is assumed to be perfectly plastic, so the corresponding reset map will result in a instantaneous velocity change for some of the coordinates. The method of calculating the reset map, $\Delta_{e}$, is based on the work in \cite{GCAS10}, and is defined by the equation
\begin{equation}\label{eq:impact}
    \dot{q}^{+}=\Delta_{e}\dot{q}^{-}=[I-((M^{-1}J_{v+}^{T})/(J_{v+}M^{-1}J_{v+}^{T}))J_{v+}]\dot{q}^{-}
\end{equation}
where $I$ is the $3x3$ identity matrix, $J_{v+}$ is the Jacobian of the constraint for the upcoming domain, and $\dot{q}^{-}$ and $\dot{q}^{+}$ represent the coordinate velocities before and after the reset, respectively. The result of this reset mapping for ground contact is the foot velocity will be immediately set to zero after the impact, while that for the mover hardstop is based on conservation of momentum principles for the mover and body so they have the same velocity after impact.
\section{Simulation}
\label{sec:sim}

\begin{figure}[t!]
	\centering
	\includegraphics[width=0.99\columnwidth]{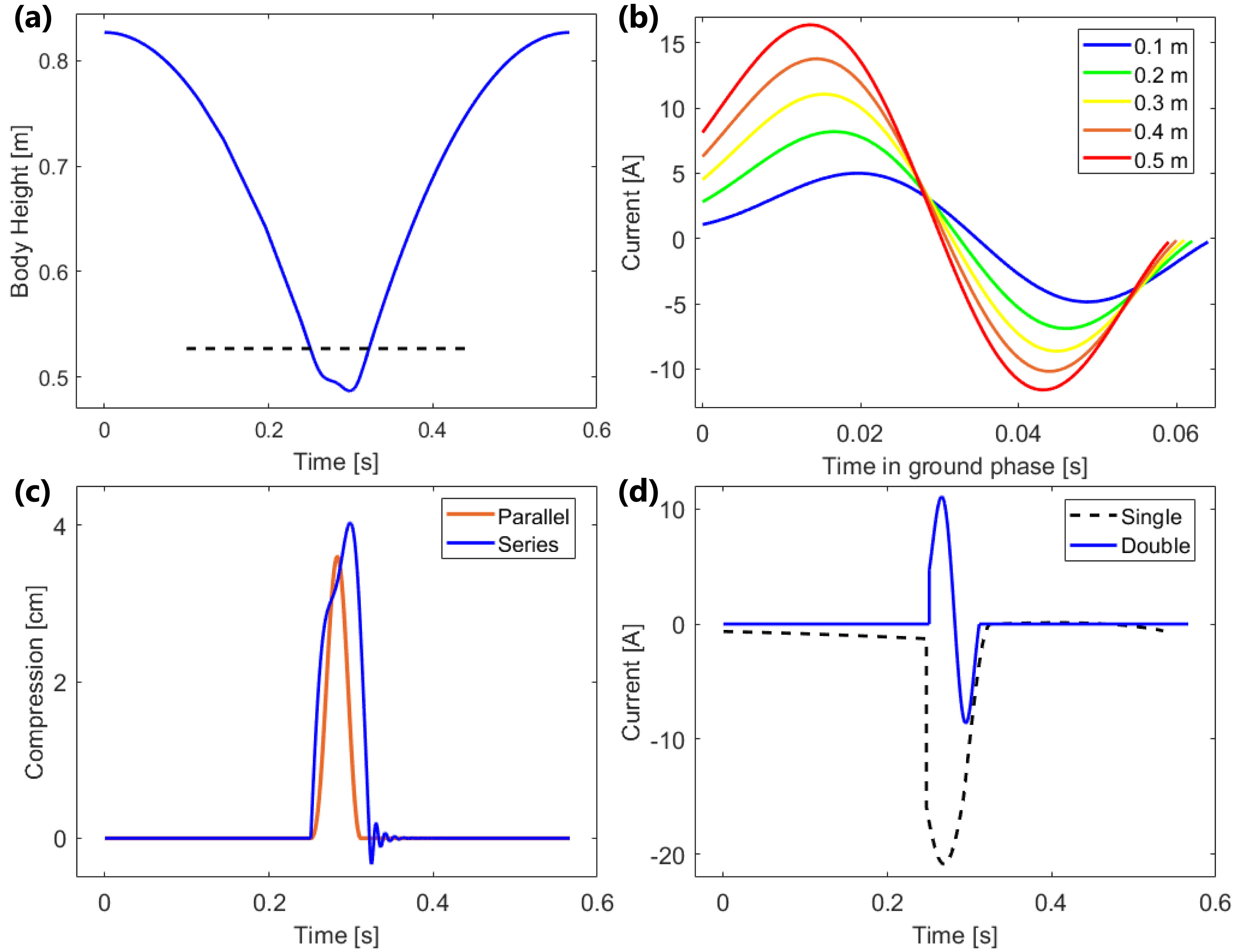}
	\vspace{-6pt}
	\setlength\belowcaptionskip{0pt}
	\caption{(a) Optimal trajectory for the body height with ground contact threshold. (b) Input current profiles for $5$ hop heights. (c) Optimal trajectory for both spring deflections ($y$ and $\delta$). (d) Input force comparison for both models.}
	\label{opt3}
\end{figure}

\subsection{Motion Generation}
Due to the complexity of hybrid system models---and thereof hopping---that combines discrete and continuous dynamics, motion generation becomes non-trivial. To this end, trajectory optimization is used to find the ideal actuation input to follow. The method of optimization utilized here is similar to that used in \cite{vision19}, with the nonlinear programming formulated as
\begin{align}
    \min_{q(t),u(t)} \quad  &\int_{t_{0}}^{t_{f}} u^{2}(t) dt \\
    s.t. \quad &\mathbf{C_{1}.}\ pinned\ dynamics      \nonumber \\
               &\mathbf{C_{2}.}\ hybrid\ continuity    \nonumber \\
               &\mathbf{C_{3}.}\ physical\ feasibility \nonumber
\end{align}
where constraint $\mathbf{C_{1}}$ is the dynamics from (\ref{eq:dyn}) and (\ref{eq:Jacobian}), $\mathbf{C_{2}}$ represents the continuity of states across each reset map, and $\mathbf{C_{3}}$ includes torque and range of motion limits present in the hardware to be used in experiment. The optimization package used for this work was \textit{GPOPS-II} \cite{GPOPS2}, which is MATLAB based and allows for multiple domains and discrete event as are required for these hopping systems.

\begin{figure}[t!]
	\centering
	\includegraphics[width=0.99\columnwidth]{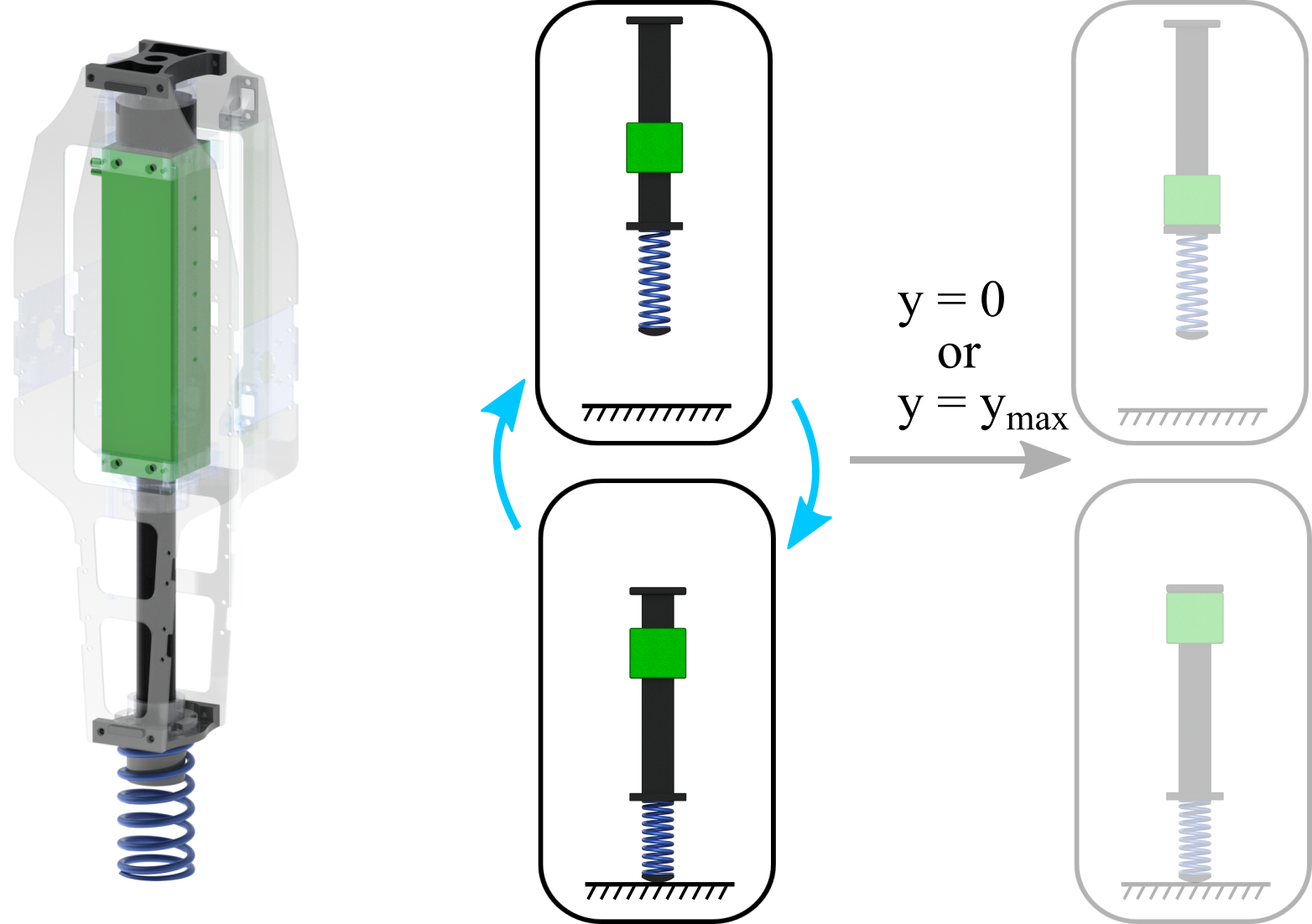}
	\vspace{-6pt}
	\setlength\belowcaptionskip{4pt}
	\caption{The model for the single-spring hopper and its hybrid system, with the possible other domains that were considered and found sub-optimal shown as grayed-out. }
	\label{model2}
\end{figure}

From previous work with the single-spring model, only two domains were planned in optimization. This was assuming that the mover would never contact the body, restricting the hybrid structure to only be based on foot contact with the ground. Alternative hybrid domain cycles were explored, allowing for this internal contact and various orders of events; see Fig. \ref{model2}. However, it was found that the optimal choice was in fact to restrict the mover to never reach contact with the body, and again to only use the two-domain structure as done previously. The optimization was re-run for the single-spring model using parameters consistent with the robot to be used in experiment, but with the absence of the second spring and its resulting damping on the motion of the mover.

For the double-spring case, multiple hybrid cycles were examined to find the optimal discrete sequence of events, i.e., the optimal graph $\Gamma$ to use in the hybrid system model. The best hybrid structure was found with stiffness constants of the two springs being close together and then skipping over domain $2$, following the path of $D_{1} \rightarrow D_{3} \rightarrow D_{4} \rightarrow D_{1}$. The process of going straight from $D_{1}$ to $D_{3}$ involves applying no input from the actuator until the foot reaches the ground, and then making sure that the initial input force is non-negative. This aligned with the preferred direction of input during the ground phase, requiring no extra constraint in the optimization. Note that in this particular hybrid cycle, there is only one domain when the mover is not locked to the body and when actuation will occur: domain $3$.

This optimization was run for multiple hop heights with foot clearance ranging from \SI{0.1}{\meter} to \SI{0.5}{\meter}. Fig. 4(a) shows the optimal trajectory of the body height coordinate for a foot clearance of \SI{0.3}{\meter}, while Fig. 4(c) shows the corresponding optimal spring deflection trajectories. Fig. 4(b) shows the optimal input force trajectories for all $5$ hop heights over the time interval of domain 3, when the upper spring is not locked in contact with the body. Fig. 4(d) shows a comparison between the optimal force profiles of the single-spring and double-spring models, both for a hop with foot clearance of \SI{0.3}{\meter}. It can be seen that the optimal force trajectory for the double-spring model follows a path similar to that of the spring in parallel, assisting its compression and extension.

\begin{table}[t!]
    \centering
    \begin{tabular}{c||ccc|ccc}
        \hline
		 & & & & & & \\ [-0.9em]
         & \multicolumn{3}{c}{Single-Spring} & \multicolumn{3}{c}{Double-Spring} \\
         \hline
         $H_{f}$ & $F_{max}$ & $\eta_{MECH}$ & $\eta_{ELEC}$ & $F_{max}$ & $\eta_{MECH}$ & $\eta_{ELEC}$ \\
         \hline
		 & & & & & & \\ [-0.9em]
         0.1 & 95.2 & 25\% & 17\% & 50.9 & 78\% & 58\% \\
         0.2 & 183.4 & 26\% & 16\% & 98.3 & 77\% & 55\% \\
         0.3 & 250.1 & 29\% & 16\% & 132.6 & 76\% & 54\% \\
         0.4 & 303.0 & 30\% & 15\% & 165.2 & 76\% & 52\% \\
         0.5 & 348.1 & 30\% & 15\% & 196.4 & 76\% & 51\% \\
         \hline
    \end{tabular}
    \caption{Key comparison data for the optimization results.}
    \vspace{6pt}
    \label{tab:comp}
\end{table}

Table \ref{tab:comp} shows further comparisons between both versions of the moving-mass model in these optimizations. Specifically, it compares peak force, mechanical efficiency, and electrical efficiency. Mechanical energy used in each hop is a function of motor force and mover velocity given by
\begin{equation}
    E_{MECH}=\int_{t_{0}}^{t_{f}} |u\dot{y}|dt
\end{equation}
Current and voltage were estimated as functions of the force and velocity of the motor using the measured resistance and back emf of the motor. Using these values of current and voltage, the total electrical energy provided in each hop was
\begin{equation}
    E_{ELEC}=\int_{t_{0}}^{t_{f}} |IV|dt
\end{equation}
where $I$ is the current in the motor, and $V$ is the voltage being sent to the motor from the motor controller. These energy consumptions were compared to the amount of energy lost in the system when the actuator provided zero force throughout an entire hop, in order to get the energetic efficiencies.

It was found that the optimal input of the double-spring model required $40\%$ less peak force than the single-spring model, due to the fact that the motor only assisted the natural motion of the upper spring rather than manipulate the mover completely on its own. The double-spring model also had about $2.5x$ better mechanical efficiency and $3x$ better electrical efficiency, despite the fact that the additional spring resulted in higher damping between the mover and body and, therefore, slightly more inherent energy loss during each hop.

\begin{figure}[t!]
	\centering
	\includegraphics[width=0.95\columnwidth]{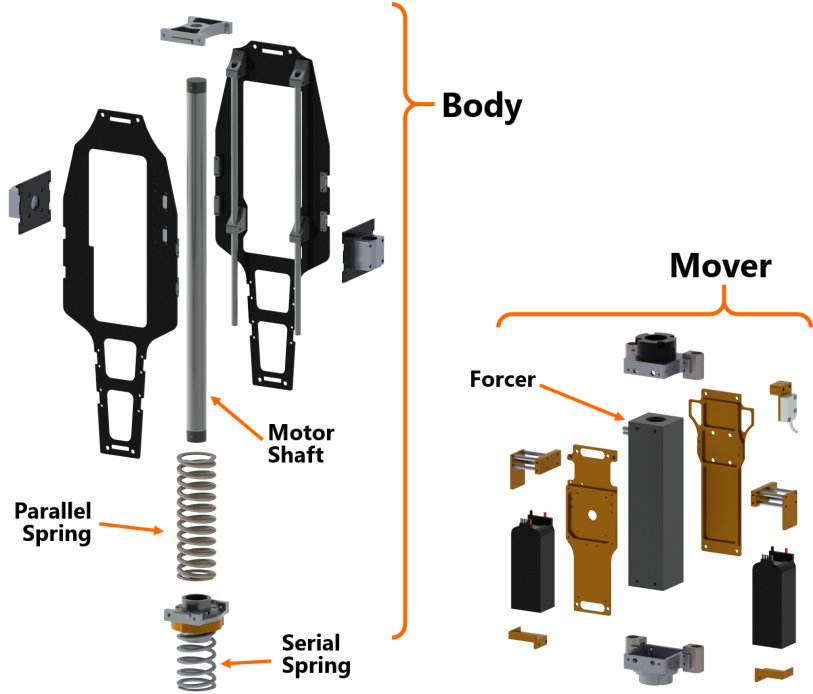}
	\setlength\belowcaptionskip{4pt}
	\caption{An exploded view of the CAD model of the robot, calling out the key internal components.}
	\label{exploded}
\end{figure}

\subsection{Simulated Hopping}
Hopping was simulated for both models in MATLAB starting with open-loop playback of the input trajectories. At the onset of every hop, time is reset and the trajectory starts again from the beginning in order to observe stability with minimal state feedback. Poincar\'e map analysis \cite{periodicHop} is used to provide a metric to the stability, where the system is considered stable if $|\lambda_{i}|<1$ for all eigenvalues of the Poincar\'e section Jacobian. Just as discovered in \cite{iros19}, the single-spring model with updated parameters is not stable using this open-loop plus time reset control method alone. For example, in the case of hopping \SI{0.3}{\meter} off the ground, the maximum eigenvalue magnitude was $\lambda_{max} = 1.711$. It is possible to achieve stability by adding a closed-loop controller around the $y$ coordinate; PD control is used in this example. By closing the loop around this coordinate, the max eigenvalue magnitude becomes $\lambda_{max} = 0.332$, allowing the robot to stabilize from errors in initial conditions.

In contrast, the double-spring model is stable through only open-loop playback and time reset. The eigenvalue magnitude for the same example hop height is $\lambda_{max} = 0.803$, without the use of a stabilizing controller. This is due to the spring in parallel forcing the $y$ coordinate along a path very similar to that desired, since the the optimal trajectory acts only to assist this deflection. Any small deviations from the optimal path are guided back to the trajectory by the spring's natural dynamics. Furthermore, the fact that control input is only occurring during a single phase lasting roughly \SI{60}{\milli\second} also limits the negative effects of any error as well.              
\section{Experimental Results}
\label{sec:exp}

\begin{figure}[t!]
	\centering
	\includegraphics[width=0.99\columnwidth]{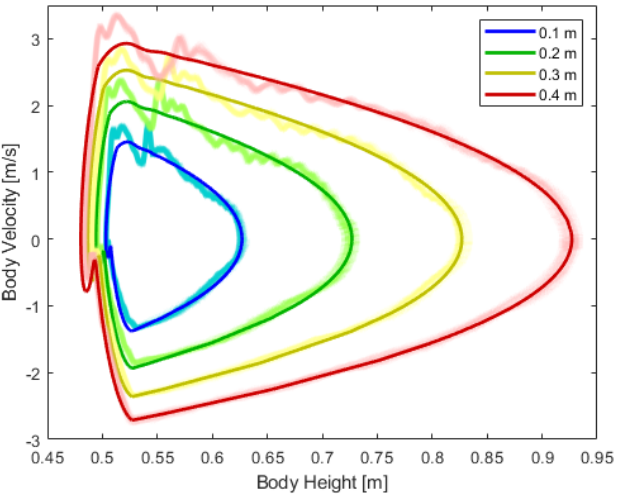}
	\vspace{-8pt}
	\setlength\belowcaptionskip{4pt}
	\caption{Phase portraits of the $z_{b}$ coordinate during 20 consecutive hops in experiment for each height. The dark curves represents the ideal trajectories from optimization, and the light curves are the experimental sensor data.}
	\label{phase}
\end{figure}

\begin{figure*}[t!]
	\centering
	\includegraphics[width=0.98\textwidth]{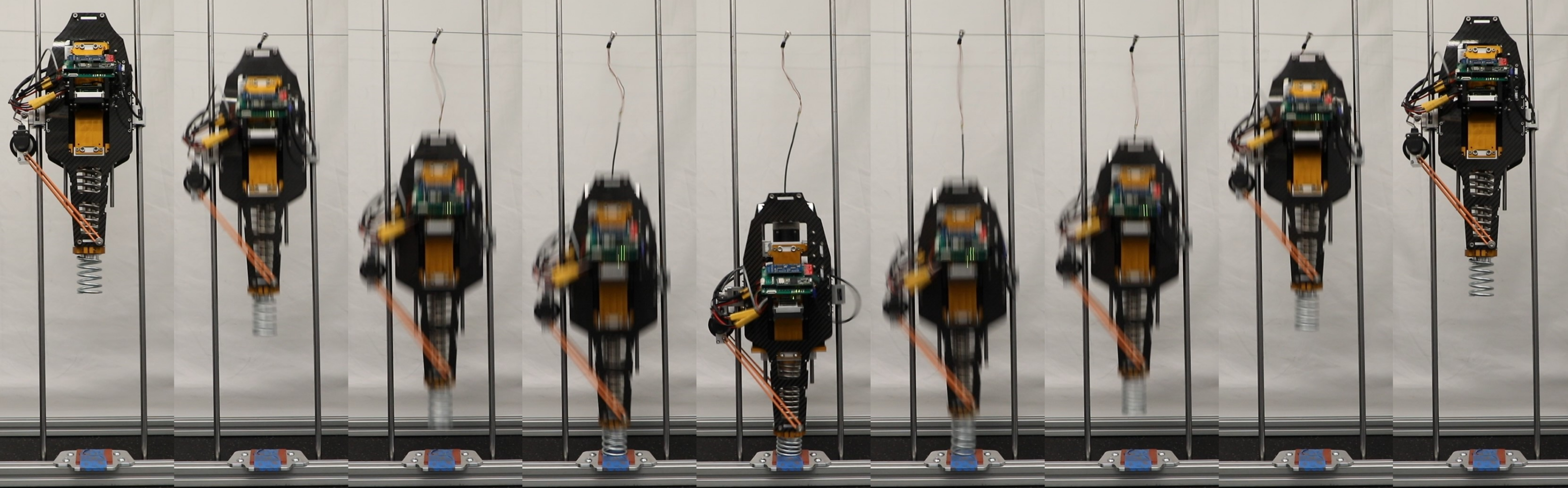}
	\setlength\belowcaptionskip{-6pt}
	\caption{Tiles from the \SI{0.3}{\meter} experiment over a single hop.}
	\label{tiles}
\end{figure*}

\subsection{Hopping Robot Hardware}
The one-dimensional hopping robot shown in Fig. \ref{pixar} was developed to demonstrate these concepts on real hardware. The robot is fixed to the vertical axis through the use of two linear rails and linear ball bearings on either side of the robot. The core of the system is built around a linear brushless DC motor from Nippon Pulse \cite{url:nippon}, which is capable of providing up to \SI{450}{\newton} of peak force and \SI{75}{\newton} of continuous force. The motor is made up of two parts: a forcer, which is a box shaped `stator', and a shaft, which acts as the `rotor'. As shown in Fig. \ref{exploded}, the body of the hopping robot is built around the motor shaft, accounting for the majority of the robot's \SI{52}{\centi\meter} height. The mover is connected to the forcer of the motor and also contains most of the robot's electronics, including the batteries, sensors, motor controller, and main processing board. The upper spring is placed around the motor shaft to act in parallel with the actuation. In order to minimize the mass of the foot, the lower spring is only held at the top using a clamp mechanism while the lower end is unconstrained. This reduces the amount of energy lost during each impact with the ground, but allows for some off-axis deflection in the spring. Parameter estimation for the system was performed with drop tests of the robot, while the motor was commanded to maintain zero current. The data from these tests was used to identify the three damping coefficients and two spring stiffness constants in the model. While estimates of the stiffness constants were provided by the manufacturer, it was found that the actual values differed slightly and were thus updated accordingly.

\subsection{Experiments}
Experiments were performed with this robot using the force input trajectories gathered from optimization for hopping motions with foot clearances ranging from \SI{0.1}{\meter} to \SI{0.4}{\meter}. As done in the simulations for this model, the robot attempted to playback the open-loop trajectory of current to the motor while on the ground. From the initial tests, it was seen that the robot was not returning to its starting height after the impacts with the ground. It was found that this was due to the internal impact between the mover and body not being perfectly plastic. Instead there was a some bouncing happening at that domain transition leading to less uniform energy transfer. This was corrected by having the motor apply a negative current briefly after the end of the trajectory to `lock' the body and mover together. With this change, the robot was able to hop back up to its intended height, with noticeably less bouncing during this impact.

Multiple experiments were run at four different heights corresponding to foot clearances of \SI{0.1}{\meter}, \SI{0.2}{\meter}, \SI{0.3}{\meter}, and \SI{0.4}{\meter} (experiments can be seen in the supplemental video [1]). A phase portrait of the experiments for all four hop heights is provided in Fig. \ref{phase}, showing the periodic nature of the hopping motion. The data further confirms the improved stability inherent to this hopping model as seen by the very consistent path from the robot during the experiments. However, it can be seen that there is significant deviation from the optimal trajectory between the time of the internal impact until the time just after the take-off of the robot. This deviation causes oscillations around the ideal path and is eventually damped out by the time the robot reaches its apex. This is partially due to the way the robot is attached to the rails in experiment and the fact that the lower spring can laterally deflect while on the ground. The linear bearings are able to rotate slightly from the rails due to backlash, so the whole robot is able to rock back and forth marginally, leading to oscillations in the body velocity during ascent. This phenomenon can also be seen in the provided video.             
\section{Conclusion}
\label{sec:conclusion}

A new double-spring version of the moving-mass hopper model was presented and compared with the previously used single-spring model. Trajectory optimization was used to identify the optimal actuation for each model, showing two different patterns of force input. The single-spring case had large force impulses during the ground phase acting in time with the spring in series. Conversely, in the double-spring case, the actuator acted in time with the natural compression of the upper spring in parallel. This led to a prolonged deflection of the spring in series, but an overall more condensed period of actuation. It was shown in simulation that adding this spring in parallel improved hopping performance through increasing energy efficiency by more than $2.5x$, reducing peak actuator effort by roughly 40\%, and providing stability to the hopping motion without the need for closed-loop control.

This model was further demonstrated on hardware using a novel vertically-constrained hopping robot designed and built in the lab. Applying the reference motor current trajectories generated in optimization, the robot was able to reach hop heights up to \SI{40}{\centi\meter} off the ground. This max height was only limited by the length of the lower spring, which was reaching its solid length at higher hop heights. In the future, an alternative elastic component will be made to replace this spring, allowing for larger deflections and increased energy storage. Additional actuators will also be added to provide the balancing mechanisms necessary for hopping in 3-dimensional space.              

\newpage
\nocite{url:video}
\bibliography{hopper_references}{}
\bibliographystyle{plain}

\end{document}